\documentclass[11pt]{article}
\usepackage{coling2018}
\usepackage{times}
\usepackage{url}
\usepackage{latexsym}
\usepackage{amsmath}
\usepackage{multirow}
\usepackage{color}
\usepackage{CJK}
\usepackage{amsfonts}
\usepackage{url}
\usepackage{graphicx}  
\usepackage{subfigure}
\usepackage{url}

\title{One-shot Learning for Question-Answering in Gaokao History Challenge}

\author{Zhuosheng Zhang$^{1,2,}$, Hai Zhao$^{1,2,}$\thanks{$\ $ Corresponding author. This paper was partially supported by
		National Key Research and Development Program of China (No. 2017YFB0304100),
		National Natural Science Foundation of China (No. 61672343 and No. 61733011),
		Key Project of National Society Science Foundation of China (No. 15-ZDA041),
		The Art and Science Interdisciplinary Funds of Shanghai Jiao Tong University (No. 14JCRZ04).}\\
	$^{1}$Department of Computer Science and Engineering, Shanghai Jiao Tong University \\
	$^{2}$Key Laboratory of Shanghai Education Commission for Intelligent Interaction \\ and Cognitive Engineering, Shanghai Jiao Tong University, Shanghai, 200240, China\\
	{\tt zhangzs@sjtu.edu.cn, zhaohai@cs.sjtu.edu.cn}
}

\date{}

\begin{document}
\maketitle
\begin{abstract}
	Answering questions from university admission exams (Gaokao in Chinese) is a challenging AI task since it requires effective representation to capture complicated semantic relations between questions and answers. In this work, we propose a hybrid neural model for deep question-answering task from history examinations. Our model employs a cooperative gated neural network to retrieve answers with the assistance of extra labels given by a neural turing machine labeler. Empirical study shows that the labeler works well with only a small training dataset and the gated mechanism is good at fetching the semantic representation of lengthy answers. Experiments on question answering demonstrate the proposed model obtains substantial performance gains over various neural model baselines in terms of multiple evaluation metrics.
\end{abstract}

\section{Introduction}
\blfootnote{
	%
	%
	%
	%
	%
	%
	This work is licensed under a Creative Commons 
	Attribution 4.0 International License.
	License details:
	\url{http://creativecommons.org/licenses/by/4.0/}
}

Teaching a machine to pass admission exams is a hot AI challenge which has aroused a growing number of research \cite{E17-1011,chengtaking,Clark2015Elementary,fujita2014overview,Henaff2016Tracking}. Among these studies, deep question-answering (QA) task \cite{ferrucci2010building,wang2014overview} is especially difficult, aiming to answer complex questions via deep feature learning from multi-source datasets. Recently, a highly challenging deep QA task is from the Chinese University Admission Examination (\emph{Gaokao} in Chinese), which is a national-wide standard examination for all senior middle school students in China and has been known as its large scale and strictness. This work focuses on \emph{comprehensive question-answering} in Gaokao history exams as shown in Table \ref{History Question Examples}\footnote{These examples are chosen as they are short, the average number of words in questions and answers are 22 and 47.}, which accounts for the major proportion in total scoring and is extremely difficult in the exams. \cite{chengtaking} made a preliminary attempt to take up the Gaokao challenge, trying to solve \emph{multiple-choice questions} via retrieving and ranking evidence from \emph{Wikipedia} articles to determine the right choices. Differently, this task is to solve comprehensive questions and has to be based on knowledge representation and semantic computation rather than word form matching in the previous work.

Although deep learning methods shine at various natural language processing tasks \cite{Wang2015Bilingual,Zhang2016Probabilistic,Qin2017Adversarial,Cai2018Seq,zhang2018NHD,Huang2018Moon,zhang2018SubMRC,He2018Syntax,zhang2018DUA,li2018Seq}, they usually rely on a large scale of dataset for effective learning. The concerned task, unfortunately, cannot receive sufficient training data under ordinary circumstances. Different from previous typical QA tasks such as community QA \cite{zhang:2016} which can enjoy the advantage of holding a very large known QA pair set, the concerned task is equal to retrieving a proper answer from textbooks organized as plain texts with guidelines of very limited number of known QA pairs. In addition, the questions are usually given in a quite indirect way to ask students to dig the exactly expected perspective of the concerned facts. If such kind of perspective fails to fall into the feature representation for either question or answer, the retrieval will hardly be successful. 

\begin{CJK*}{UTF8}{gkai}
	\begin{table*}[h]
		
		\begin{center}
			\begin{tabular}{|p{15.5cm}|}
				\hline
				{\bf Q}: \emph{“农民可能充当一种极端保守的角色，也可能充当一种具有高度革命性的角色。”试结合有关史实评析这一观点。} \\``Peasants may act as an extremely conservative role, or may be highly revolutionary." Try to analyze this view in the light of the relevant historical facts. \\ 
				{\bf A}: \emph{农民阶级的经济地位和所处的时代条件决定了它可能充当一种具有高度革命性的角色。以太平天国运动为例，在斗争过程中颁布的《天朝田亩制度》就突出反映了农民阶级要求废除封建土地所有制的强烈愿望，表现出了高度的革命性。农民阶级存在的这种两面性是由其经济地位即受地主阶级压迫和小生产者的地位所决定的。} \\The economic and social conditions of peasants determine that they may act as a highly revolutionary role. Taking the Taiping Heavenly Kingdom movement as an example, the promulgated regulation ``heavenly land system" reflected the peasant class's demands and strong desire of abolishing the system of feudal land ownership, which shows a strong degree of revolution. The dual nature of the peasant class is also the result of its economic status, that is, under the oppression of the landlord class and the status of the small producers.\\
				\hline
				{\bf Q}: \emph{有人说“文艺复兴”是一场复古运动，你如何看待？如何评价其意义？} \\Some people think the Renaissance is a retro campaign. What's your opinion and how to evaluate its historic significance?\\  
				{\bf A}: \emph{文艺复兴运动从表面的含义来看，是一种复兴古希腊罗马时期的哲学、文学和艺术的活动，是一种复古运动，但从其深层的含义看，它却是一场思想解放运动，是一次思想领域里的变革。}  \\The Renaissance is seemingly regarded as a retro movement of philosophy, literature and art of ancient Greece and Rome. However, it is actually an ideological liberation movement and a revolution of thoughts from a deeper inspection.\\
				\hline

			\end{tabular}
		\end{center}
		\caption{\label{History Question Examples} Comprehensive question examples from Gaokao history exams. }
	\end{table*}
\end{CJK*}

Generally speaking, for the Gaokao challenge, knowledge sources are extensive and no sufficient structured dataset is available, while most existing work on knowledge representation focused on structured and semi-structured types \cite{Khot2017Answering,khashabi2016question,Vieira:2016}. With regard to the answer retrieval, most current research focused on the factoid QA \cite{Dai:2016,Yin:23}, community QA \cite{zhang2017attentive,lu2017community} and episodic QA \cite{Samothrakis2017Convolutional,xiong:2016,Vinyals:2016}. Compared with these existing studies, the concerned task is more compositive and more comprehensive and has to be done from unstructured knowledge sources like textbooks. Moreover, the answers of our Gaokao challenge QA task are always a group of detail-riddled sentences with various lengths rather than merely short sentences as focused on in previous work \cite{Yin:23,Yih2014Semantic}.

Recent research has turned to semi-supervised methods to leverage unlabeled texts to enhance the performance of QA tasks via deep neural networks \cite{yang2017semi,Lei:2016}. This task is somewhat different from previous ones that the expected extra labels are difficult to be annotated and the entire unlabeled data is kept in a very small scale, so that semi-supervised methods cannot be conveniently applied. Notably, one-shot learning has been proved effective for image recognition with few samples \cite{Li:2006}, which is a strategy similar to people learning concept. As an implementation of one-shot learning, neural turing machine (NTM) was proposed \cite{Santoro:2016,Vinyals:2016} and showed great potential by learning effective features from a small amount of data, which caters to our mission requirements. Inspired by the latest advance of one-shot learning, we train a weakly supervised classifier to annotate salient labels for questions using a small amount of examples. For question answering, we propose a cooperative gated neural network (CGNN) to learn the semantic representation and corresponding relations between questions and answers.  We also release a Chinese comprehensive deep question answering dataset to facilitate the research.

The rest of this paper is organized as follows. The next section introduce our models, including retrieval model, CGNN, and NTM labeler. Task details and experimental results are reported in Section 3, followed by reviews of related work in Section 4 and conclusion in Section 5.

\begin{figure*}
	
	\subfigure{
		\begin{minipage}[b]{0.45\textwidth}
			\centering
			\includegraphics[width=1\textwidth]{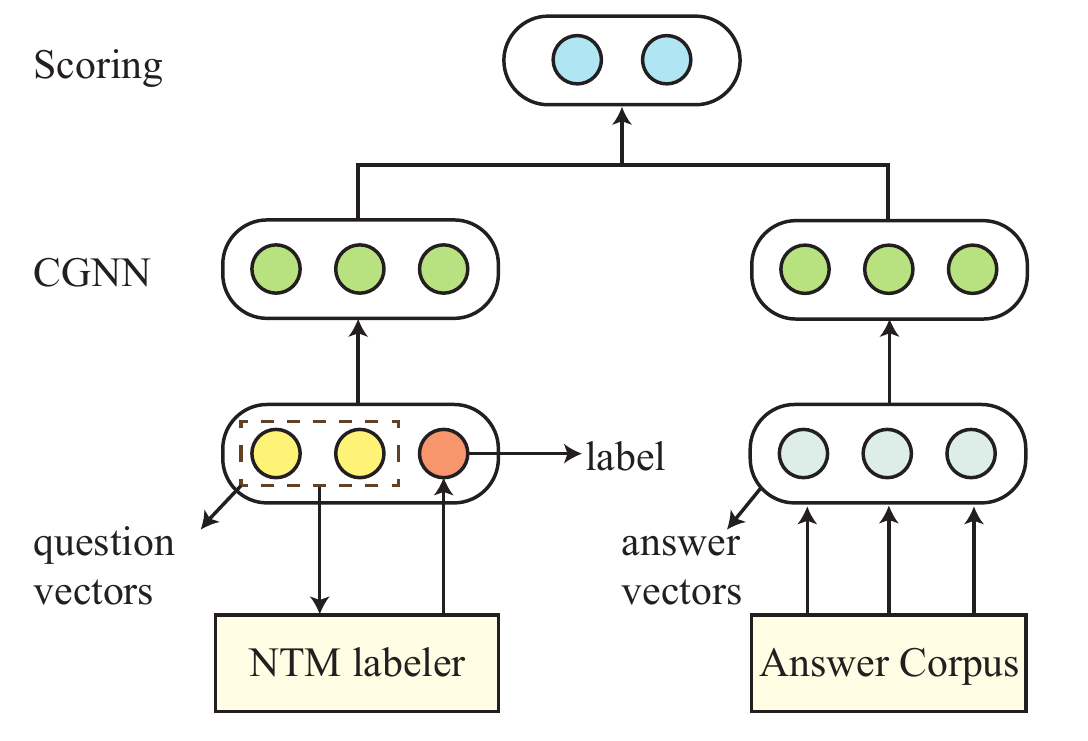}
			\caption{Model architecture.}
		\label{Fig:sys}
		\end{minipage}
	}
	\subfigure{
		\begin{minipage}[b]{0.52\textwidth}
			\centering
			\includegraphics[width=1\textwidth]{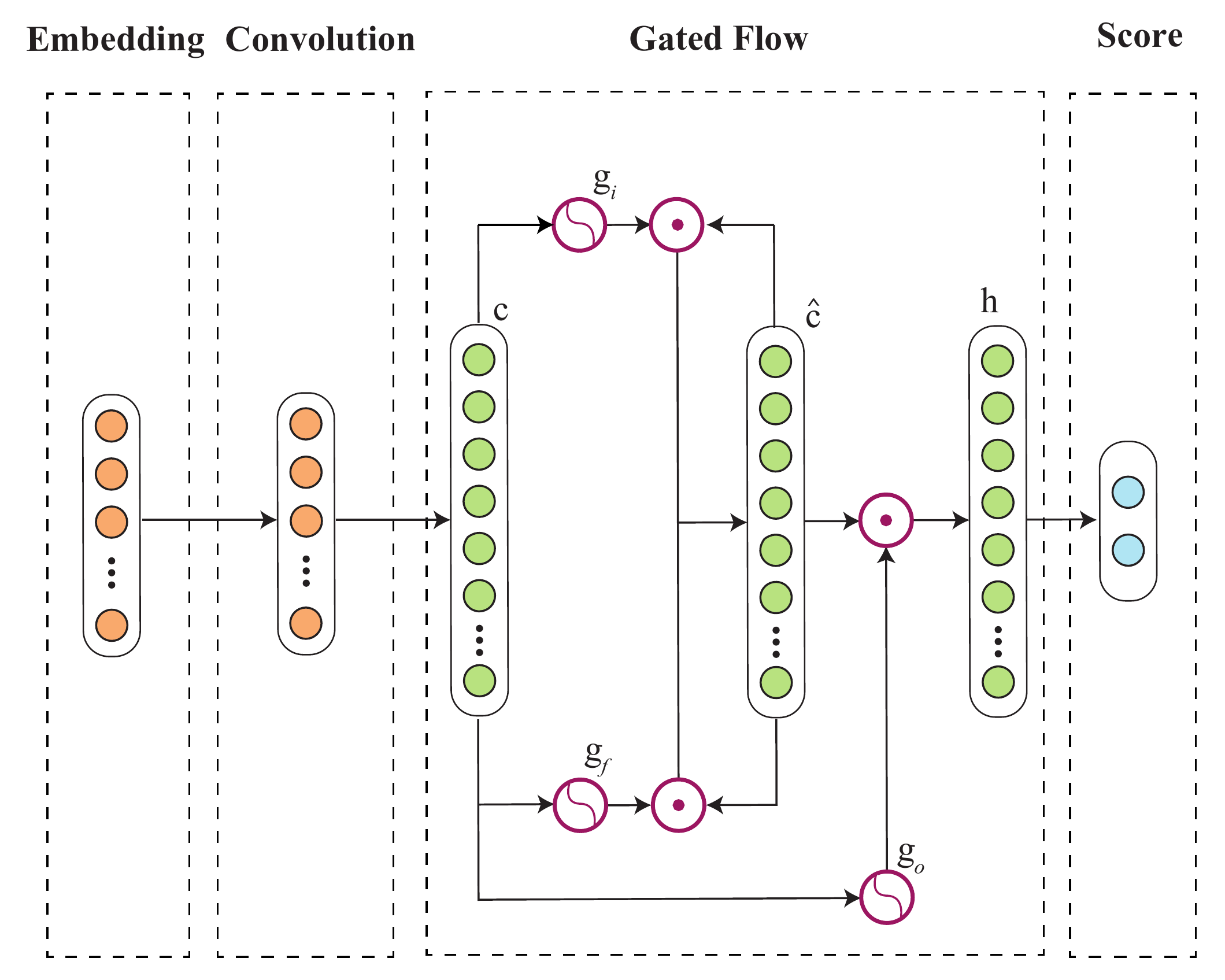}
			\caption{CGNN architecture.}
			\label{fig:CGNN}
		\end{minipage}
	}
\end{figure*} 

\section{Model}
The proposed hybrid neural model is composed of two main parts: cooperative gated neural network (CGNN) for feature representation and answer retrieval, and NTM as one-shot learning module for extra labeling. As shown in Figure \ref{Fig:sys}, we use NTM to classify the question type and annotate the corresponding label. Then, the concatenated label vectors and question representation are fed to CGNN for jointly scoring candidate answers.

\subsection{Main framework}
For the QA task, the key problem is how to effectively capture the semantic connections between a question and the corresponding answer. However, the questions and answers are lengthy noisy, resulting in poor feature extraction. Inspired by the success of the popular gated mechanism \cite{Wang2017Gated,chen2016implicit,Lei:2016} and the gradient-easing Highway Network \cite{srivastava2015highway}, a CGNN is proposed to score the correspondence of inputted QA pairs. The architecture is shown in Figure \ref{fig:CGNN}. 

\paragraph{Embedding} 
Our model reads each word of questions and answers as the input. For an input sequence, the embedding is represented as $\mathbf{M}\in \mathbb{R}^{d\times n}$ where $d$ is dimension of word vector and $n$ is the maximum length. When using the NTM module for question type labeling, the question embedding will be refined as the concatenation of the label vectors and its original embedding. Considering the calculation efficiency, we specialize a max number of words for the input and apply truncating or zero-padding when needed.

\paragraph{Convolutional Layer}
Filter matrices [$\textbf{W}_1$, $\textbf{W}_2$, \dots , $\textbf{W}_k$] with several variable sizes  [$l_1, l_2,\dots , l_k$] are utilized to perform the  convolution operations for input embeddings. Via parameter sharing, this feature extraction procedure is kept the same for questions and answers. For the sake of simplicity, we will explain the procedure for only one embedding sequence. The embedding will be transformed to sequences $\textbf{c}_j (j \in [1, k])$ : 
\begin{align*}
\mathbf{c}_j  = [\dots; \tanh (\mathbf{W}_j\cdot \mathbf{M}_{[i:i+l_j-1]} + \mathbf{b}_j); \dots]
\end{align*}
where $[i:i+l_j-1]$ indexes the convolution window. Additionally, we apply wide convolution operation between embedding layer and filter matrices, because it ensures that all weights in the filters reach the entire sentence, including the words at the margins.

\paragraph{Gated Flow}

To highlight the key information and ignore irrelevant ones during convolution, we adopt an adaptive gated decay mechanism for gated information flow. Three gates are added to optimize the feature representation. These gates are only influenced by the original input through different parameters.  Let $\hat{\mathbf{c}}_{j}^{n}$ denote $n$-gram features in $c_j$, the gates are formulated as
\begin{align*}
&\mathbf{g}_{i}=\sigma (\mathbf{W}_{i} \hat{\mathbf{c}}_{j}^{n}+\mathbf{b}_{i})\\
&\mathbf{g}_{f}=\sigma (\mathbf{W}_{f} \hat{\mathbf{c}}_{j}^{n}+\mathbf{b}_{f})\\
&\mathbf{g}_{o}=\sigma (\mathbf{W}_{o} \hat{\mathbf{c}}_{j}^{n}+\mathbf{b}_{o})
\end{align*} 
where $\sigma$ denotes sigmoid function which guarantees the values in the gates are in [0,1] and $\mathbf{W}_{i}$, $\mathbf{W}_{f}$, $\mathbf{W}_{o}$, $\mathbf{b}_{i}$, $\mathbf{b}_{f}$, $\mathbf{b}_{o}$ are model parameters.  The transformed inner cell is iteratively calculated by:
\begin{align*}
\hat{\mathbf{c}}_{j}^{n}=\hat{\mathbf{c}}_{j-1}^{n}\odot \mathbf{g}_{i}+\mathbf{g}_{f}\odot (\hat{\mathbf{c}}_{j-1}^{n-1}+\mathbf{W}_{n})
\end{align*}
where $\odot$ denotes element-wise multiplication and $\mathbf{W}^{o}$ is the parameter. Through gated flow, the vector $\mathbf{c}_{t}$ captures the filtered information. The output of our CGNN is $\mathbf{h}_{j} = \tanh(\hat{\mathbf{c}}_{j}^{n})\odot\mathbf{g}_{o}$. The gates are used to route information through the flow. Although these gates are generated independently, they work cooperatively since they jointly control the information flow of inner-cells. This procedure will help select salient features. When there is one gate in our network, the model works similarly to the original highway network \cite{srivastava2015highway}.

A \emph{one-max-pooling} operation is adopted after the gated flow and the current state vector $\mathbf{s}_{j}$ is obtained through concatenating all the mappings for those $k$ filters, we have $\mathbf{s}_{j} = \mathbf{max}(\mathbf{h}_j) $.

For discriminative training, we use a max-margin framework for learning (or fine-tuning) parameters $\theta$. Specifically, a scoring function $\varphi(\cdot,\cdot;\theta)$ is defined to measure the similarity between the corresponding representations of questions and answers. In this work, we use cosine similarity for the measurement. Let $p=\{p_1,...p_n\}$ denote the answer corpus and $a \in p$ is the answer to the question $q_{i}$, the optimal parameters $\theta$ are learned by minimizing the max-margin loss:  
\begin{align*}
\max \{\varphi(q_{i},p_{i};\theta)-\varphi(q_{i},a;\theta)+\delta(p_{i},a)\}
\end{align*}
where $\delta(.,.)$ denotes a non-negative margin and $\delta(p_{i},a)$ is a small constant when $a\neq p_{i}$ and 0 otherwise. 

\subsection{Question Type Labeling}
\begin{figure}
	\centering
	\includegraphics[width=0.75\textwidth]{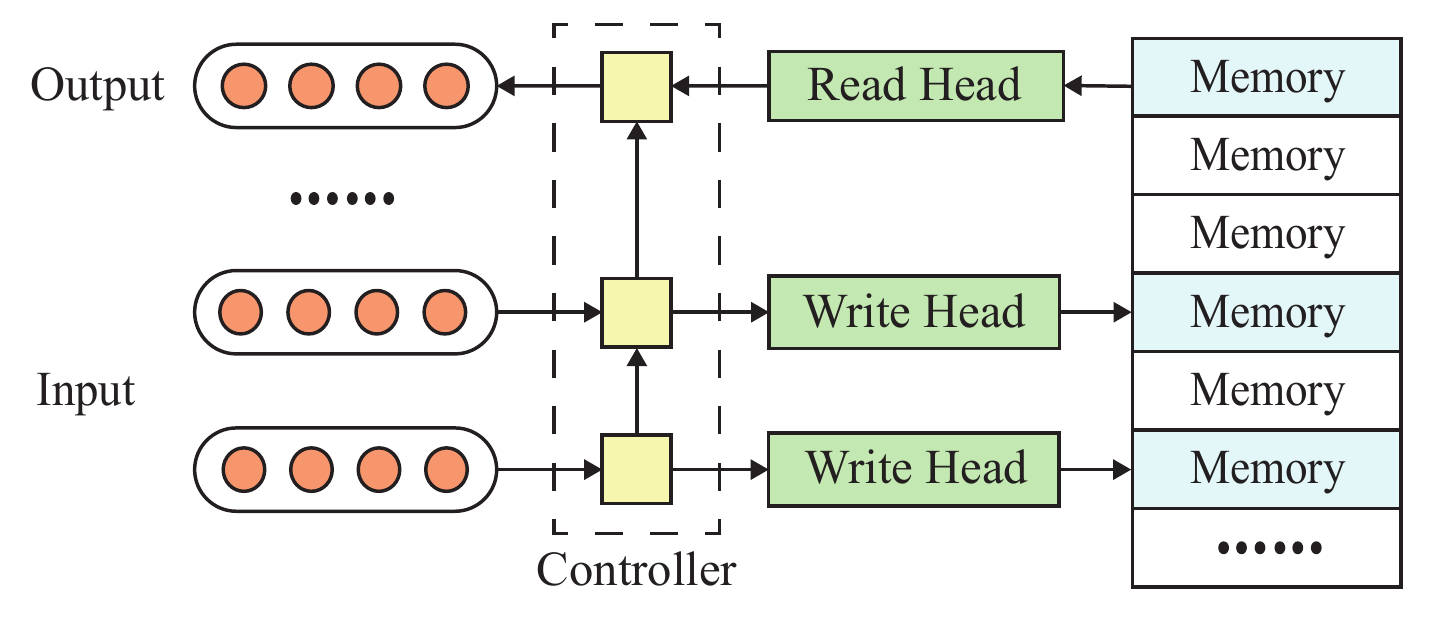}
	\caption{NTM architecture}
	\label{fig:NTM}
\end{figure}

Since examinees studying for exams are required to understand the history knowledge from different sides, we exactly consider using especially-annotated labels as extra indicators for machines to capture such features. With external memory, NTM has shown great potential for one-shot learning \cite{Santoro:2016}. As in Figure \ref{fig:NTM}, an NTM consists of two primary components, controller and memory. The controller, often implemented as a recurrent neural network, interacts with an external memory module using soft \emph{read heads} to retrieve representation from memory and \emph{write heads} to store memory, respectively.
\paragraph{Controller}
The controller of NTMs can be implemented as either a recurrent neural network or a feedforward neural network. Our model adopts LSTM cells as the implementation for the best performance \cite{Graves2014Neural}.
\begin{align*}
&\mathbf{i_{t}}=\sigma (\mathbf{W}_{w}^{i}\mathbf{w}_{t}+\mathbf{W}_{h}^{i}\mathbf{h}_{t-1}+\mathbf{b}_{i})\\
&\mathbf{f_{t}}=\sigma (\mathbf{W}_{w}^{f}\mathbf{w}_{t}+\mathbf{W}_{h}^{f}\mathbf{h}_{t-1}+\mathbf{b}_{f})\\
&\mathbf{u_{t}}=\sigma (\mathbf{W}_{w}^{u}\mathbf{w}_{t}+\mathbf{W}_{h}^{u}\mathbf{h}_{t-1}+\mathbf{b}_{u})\\
&\mathbf{c_{t}}=\mathbf{f}_{t}\odot \mathbf{c}_{t-1}+\mathbf{i}_{t}\odot\tanh(\mathbf{W}_{w}^{c}\mathbf{w}_{t}+\mathbf{W}_{h}^{c}\mathbf{h}_{t-1}+\mathbf{b}_{c})\\
&\mathbf{h_{t}}=\tanh (\mathbf{c}_{t})\odot \mathbf{u}_{t}\\
&\mathbf{o_{t}}=(\mathbf{h}_{t} \oplus \mathbf{m}_{t})
\end{align*}
where $ \oplus $ denotes vector concatenation, and $\mathbf{i_{t}},\mathbf{f_{t}},\mathbf{u_{t}},\mathbf{c_{t}},\mathbf{h_{t}}$ are the input gates, forget gates, output gates, cell state and hidden state, respectively. 

Given the input sequence, the controller computes the concatenated state sequence $\mathbf{o}_t$ by applying the formulation for each time step. Each word of the sequence is represented as word embedding. In order to simplify the calculation, we apply \emph{one-max-pooling} to each input. As a result, each sequence is represented as a vector with the same dimension of word embedding.
\paragraph{Memory Manipulation}
We use a rectangular matrix $M\in \mathbb{R}^{N\times M}$ to denote the memory module. 
Given the read vector $\mathbf{r}_t$, the memory $\mathbf{m}_t$ is retrieved by:
\begin{align*}
\mathbf{m}_{t} = \mathbf{r}_{t}\mathbf{M}_{t}, \mathbf{r}_{t} = \frac{\textrm{exp}(\mathbf{K}(\mathbf{k}_{t},\mathbf{M}_{t}))}{\sum_{j}\textrm{exp}(\mathbf{K}(\mathbf{k}_{t},\mathbf{M}_{t})}
\end{align*}
where $\mathbf{r}_{t}$ is read vector and $\mathbf{K}$ is  the similarity score.
For writing, the memory is updated by 
\begin{align*}
\mathbf{M}_{t} = \mathbf{M}_{t-1}\cdot (1-\mathbf{w}_{t}\mathbf{e}_{t})+\mathbf{w}_{t}\mathbf{c}_{t}
\end{align*}
where $\mathbf{w}_{t}$, $\mathbf{e}_{t}$ and $\mathbf{c}_{t}$ represent the write vectors, erase vectors and content vectors respectively.

There are two categories for memory addressing mechanism of the original NTM: content-based addressing comparing each memory locations by some key $\mathbf{k}_{t}$ and a location-based addressing by shifting the heads, akin to running along a tape. However, the location-based addressing is not optimal for conjunctive coding of information independent of sequence. Following \cite{Santoro:2016}, we use an addressing strategy called the Least Recently Used Access (LRUA).
\paragraph{Addressing}
The weight $\mathbf{u}_{t}$ is defined as follows,
\begin{align*}
\mathbf{u}_{t} = \gamma \mathbf{u}_{t-1} +\mathbf{r}_{t} + \mathbf{w}_{t}
\end{align*}
where $\gamma$ is a decay parameter. The least-used vectors, $\mathbf{v}_{t}$, are boolean variables. For a given time-step, each element of $\mathbf{v}_{t}$ is set to 0 if it is greater than the smallest element of $\mathbf{u}_{t}$, otherwise is 1. The write weights are obtained by
\begin{align*}
\mathbf{w}_{t} = \sigma (g_t) \mathbf{r}_{t-1} +(1-\sigma (g_t))\mathbf{v}_{t-1}
\end{align*}
where $g_t$ is a scalar interpolation in the range (0,1) that blends the weights of previous read weights and previous least-used weights.

\paragraph{Learning}
For our labeling task, we define a cost function as the negative log-likelihood:
\begin{align*}
L(\theta)=-\sum_{n=1}^{N}{\mathbf{y}}_{n}log(Pr(\hat{\mathbf{y}}_{n}))
\end{align*}
where $\mathbf{y}_{n}$ is the label and $\hat{\mathbf{y}}_{n}$ is the predicted one. After perspective detection, the label is then fed to the CGNN module as one-hot vectors and concatenated with the question representation. 
\section{Experiments}

\paragraph{Corpus}
\begin{table}
	\centering
	{
		\begin{tabular}{lcc}
			\hline
			\hline
			& Train & Test \\
			\hline
			\# QA pairs  &  964 & 965 \\
			Avg \# words in questions & 18 & 25 \\
			Avg \# words in answers & 46 & 48 \\
			Max \# words in questions & 478 & 482 \\
			Max \# words in answers & 4,652 & 4,387 \\
			
			\# Candidate answer set & \multicolumn{2}{c}{1,929} \\
			\# Vocabulary & \multicolumn{2}{c}{10,806} \\
			\hline
			\hline
		\end{tabular}
	}
	\caption{\label{tab:dataset} Data statistics of the comprehensive question-answering corpus.}.
\end{table}

Gaokao challenge is an imitation of open-book examination for computers. Therefore, only quite a limited number of resources can be used, which makes common semi-supervised methods unlikely beneficial from large scale unlabeled data. In addition, all our source for the answer should come from standard textbook which contains unstructured plain texts as we initialized our system building.

The corpus is from the standard history textbook\footnote{Standard Middle-school History Textbook (Vol. 1-3), published by the People's Education Press in May, 2015.}. First, to compose an answer set, 1,929 text fragments\footnote{1,929 is the number of all must-to-be-mastered history facts required by national education quality control.} were extracted by human experts. Then equal numbers of questions were collected from past Gaokao exam papers, and their answers were manually assigned from the answer set to give 1,929 annotated QA pairs, which were equally split for training and test. All text fragments in either questions or answers are segmented into words using \emph{BaseSeg} \cite{Hai2006An}. We publish it to research communities to facilitate the research \footnote{Our source is available at: \url{https://github.com/cooelf/OneshotQA}.}. Data statistics are in Table \ref{tab:dataset}.

\begin{table}[h]
	\begin{center}
		\begin{tabular}{l|l|l}
			\hline
			\hline
			\multirow{2}{*}{Embedding}
			&Max number of words &$ n=100 $ \\
			&Word embedding size & $d=200$ \\
			\hline
			\multirow{4}{*}{CGNN}&
			Hidden unit number & $h=200$ \\
			&Initial learning rate &$lr=10^{-3}$\\
			&Regularization &$\nabla = 10^{-5}$ \\	
			&Dropout &$p = 0.2$ \\
			&Filter Width &$k = 4$ \\	
			\hline
			\multirow{3}{*}{NTM} &
			Controller size&$ s=200 $ \\
			&Read heads & $r=4$ \\
			&Memory shape & $m=(128,100)$ \\ 
			&Initial learning rate & $lr=10^{-3}$ \\	
			\hline 	
			\hline
		\end{tabular}
	\end{center}
	\caption{\label{UEEQAStat} Hyper-parameters of our model. }
\end{table}

 \begin{CJK}{UTF8}{gkai}
	\makeatletter\def\@captype{table}\makeatother
	\begin{center}
		\begin{tabular}{|l|p{12.5cm}|}
			\hline
			Class Labels & Answers \\
			\hline
			& 运用我们从古代诗文、戏曲、民间传说中已经学到的知识，举例说明中国古代自给自足的自然经济的状况。\\
			背景
			Background & Using the knowledge learned from ancient poetic proses, operas, and
			folklores, exemplify the situation of self-sufficiency of natural economy
			in ancient China.\\
			\hline
			& 春秋战国时期是社会剧烈动荡的历史阶段，为什么在这样的时期
			会出现思想文化活跃的局面。\\
			原因
			Cause & The Spring-autumn and Warring States Period was the historical stage of
			the violent social unrest. Why would there be an active ideological and
			cultural phenomenon in such a period? \\
			\hline
			& 在启蒙运动中，众多的启蒙思想家的共性思想主张是什么？ 他们
			之间有何继承和发展。\\
			主张
			Claim & During the Enlightenment, what is the common thought of those
			enlightening thinkers? What is the inheritance and development
			between them? \\
			\hline
			& ``农民可能充当一种极端保守的角色，也可能充当一种具有高度革
			命性的角色。" 试结合有关史实评析这一观点。\\
			事实
			Fact & ``Farmers may act as an extremely conservative role, or may be highly
			revolutionary." Try to analyze this view in the light of the relevant
			historical facts. \\
			\hline
			& 用历史唯物主义和辩证唯物主义的观点来分析古代雅典民主政治
			和罗马法发展的历程，了解它们对后世的作用和影响。\\
			意义
			Influence & From the perspective of historical and dialectical materialism, analyze
			the history of Athens's democratic politics and the development of
			Rome law in ancient times, and understand their roles and influences
			on later generations. \\
			\hline
		\end{tabular}
	\end{center}
	\caption{\label{Text Fragments} Examples of Text Fragments of 5 classes. }
\end{CJK}
\paragraph{Setup}
For the sake of computational efficiency, a maximal length of 100 words for each text fragment is specialized and truncating or zero-padding will be applied as necessary. Word embedding is trained by word2Vector \cite{mikolov:2013} using \emph{Wikipedia} corpus\footnote{\url{https://dumps.wikimedia.org/} }.

The diagonal variant of AdaGrad \cite{Duchi2010Adaptive} is used for neural network training. Table \ref{UEEQAStat} shows the hyper-parameters of our models. For other neural models, we fine-tune the hyper-parameters with the following range of values: learning rate $\in \left \{ 1e-3,1e-2 \right \}$, dropout probability $\in \left \{ 0.1, 0.2 \right \}$, CNN filter width $\in \left \{ 2, 3, 4 \right \}$. The hidden dimensions for all the neural networks are 200.

In the following experiments, we use NTM to annotate salient labels for each question. Then, the question representation and corresponding label vectors are concatenated and fed to CGNN for feature learning.

\paragraph{Labeling}


Examinees are required to understand the history knowledge from different sides. Thus, the solver should also be capable of capturing such kind of features. To specify a key perspective on historical questions, five classes\footnote{These classes are determined according to national education quality control who requires every student should clearly distinguish these specific perspectives about history facts. More details are in Supplementary Materials.} (\emph{background, cause, claim, fact and influence}) are annotated as shown in Table \ref{Text Fragments}. Note that such kind of annotation is even not easy for human annotators or history teachers and our complete set for all answers is very small (less than 2,000). At last, we have 80 answers per class annotated for a total number of 400, in which 50 are selected for training\footnote{As our labeling is a five-class classification task, 50 samples therefore allow 10-shot learning per class at most.} and 350 for test. For our detailed case, our data is too precious to be used too many for training.

One-shot learning is supposed to effectively represent the questions in the feature space with a small set of training data. To evaluate its performance, we specify a fixed number of training data. At each epoch, NTM is randomly provided with a specific number (shot) of samples per class from training set according to the typical training setting of one-shot learning \cite{Santoro:2016}. One-shot learning, in previous practice, usually still took a large candidate training set (i.e., hundreds) as input, though for each class, only a small fixed number of shots (i.e., less than 10) were randomly selected and fed for learning. However, it is quite different from our data reality and we have to keep a least training set (10 for our case) for one-shot learning.

\begin{table}[t]\centering\small
	\centering
	{
		\begin{tabular}{l|c|c|c|c}
			\hline
			\hline
			Method & 1-shot & 2-shot & 5-shot & 10-shot\\
			\hline
			Naive Bayes & 22.1 & 25.7 & 30.1 & 34.1 \\
			K-Means & 12.5 & 15.7 & 15.8 & 16.1 \\
			SVM & 19.1 & 21.8 & 30.2 &  \textbf{45.2} \\
			\hline
			Feedforward & 20.1 & 22.7 & 21.3 & 21.6 \\	
			LSTM & 23.8 & 28.5 & 32.5 & 38.2 \\
			\hline
			NTMs & \textbf{38.7} &  \textbf{49.1} &  \textbf{63.9} &  \textbf{76.6} \\
			\hline
			\hline
		\end{tabular}
	}
	\caption{\label{tab:oneshot} Accuracy for labeling task.}
\end{table}

\makeatletter\def\@captype{figure}\makeatother
\begin{minipage}{.45\textwidth}
	\centering
{
	\includegraphics[width=0.9\textwidth]{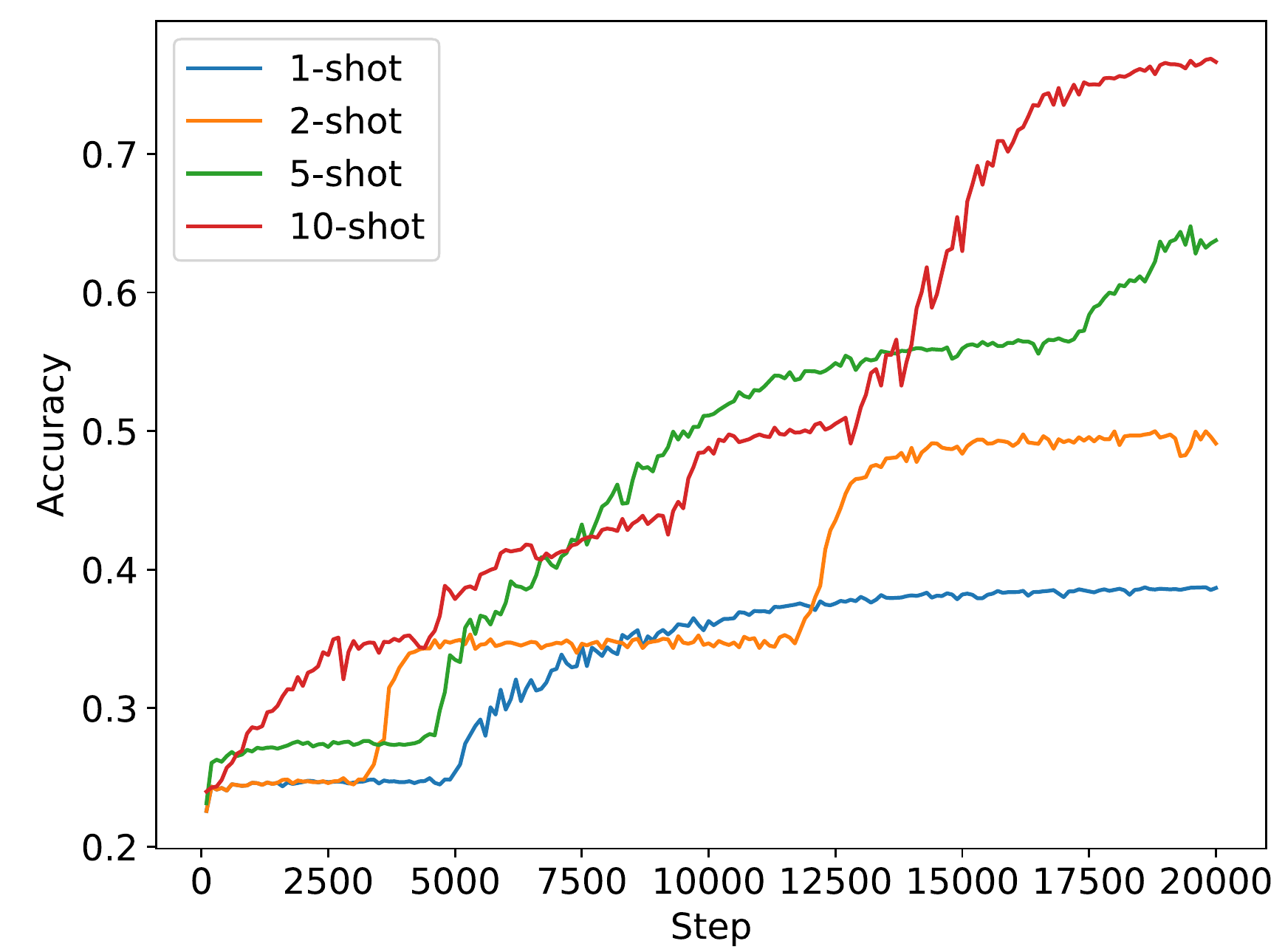}
	\caption{Labeling accuracy curve.}
	\label{fig:oneshot}
}
\end{minipage}
\makeatletter\def\@captype{table}\makeatother
\begin{minipage}{.5\textwidth}

{\centering\small
	\begin{tabular}{l|c|c|c}
		\hline
		\hline
		Method  & Precision	 & Recall & F1-score \\
		\hline
		BM25 & 8.6 (8.7) & 23.2 (23.7)  & 12.5 (12.7)\\
		\hline
		CNN & 9.8 (27.0)  & 21.1 (45.0) & 13.4 (33.8) \\
		GRU & 11.4 (32.9)  & 24.1 (50.4)& 15.5 (39.8)\\
		LSTM &  8.4 (30.6) & 21.2 (48.0) &  12.0 (37.4)\\
		+ & \textbf{22.2} & \textbf{26.8} & \textbf{25.4} \\
		\hline
		CNN+Highway & 15.8 (33.2)  & 30.5 (50.6)  & 20.8 (40.1)\\
		CNN+GRU &  17.4 (32.4) & 31.1 (50.8) & 22.3 (39.6)\\
		CNN+LSTM &  12.7 (27.5) &  21.4 (47.0) & 16.0 (34.7)\\		
		\hline
		CGNN & \textbf{18.2 (33.7)} & \textbf{31.6 (51.6)} & \textbf{23.1 (40.8)} \\
		+ & 15.5 & 20.0 & 17.7\\
		\hline
		\hline
	\end{tabular}
}
\caption{\label{tab:AR} Answer retrieval performance without and with NTM labels (in brackets). "+" means the performance gains with the help of NTM labels.}
\end{minipage}

Table \ref{tab:oneshot} compares NTM with traditional classifiers (Naive Bayes, K-Means and SVM) and neural networks (Feedforward network and LSTM). All the baseline methods are fed the training instances as the same shot style of NTM and parameters are tuned for best performance.

Figure \ref{fig:oneshot} illustrates NTM learning curves with different shots. The results show that NTM works effectively on a small amount of samples. With memory cells as well, LSTM is chosen as the baseline. The hidden layer dimension of the LSTM is 200, which is the same as the NTM controller size. The LSTMs are fed with the training instances as the same shot style of NTM. Results shown in Table \ref{tab:oneshot} indicate only NTM provides satisfactory performance with the least input instances. The effectiveness of NTM may attribute to its capability of slowly learning the representations of raw data through weight updates, and rapidly binding relevant information after a single representation via an external memory. To that extend, the NTM could generalize the meta features of each question type and distinguish the perspectives. This indicates one-shot learning is able to lessen over-fitting issues from sparse features of limited data.

\paragraph{Answer Retrieval}

For a discriminative training, we add negative QA pairs by randomly selecting 20 false answers for each positive QA pairs in the original training set. For evaluation, all answers in the corpus are ranked to match each question. NTM labeler is from 10-shot training.

We use baselines including BM25 implemented by a standard search engine \emph{Apache Lucene}\footnote{\url{http://lucene.apache.org/}} and other neural models, Convolutional Neural Network (CNN), LSTM, Gated Recurrent Unit (GRU). Our evaluation is based on the following metrics: Precision, Recall and F1-score, which are widely used for relevance evaluation. 

Table \ref{tab:AR} presents the experimental results of all the models with and without NTM labels. All of the neural networks greatly outperform BM25 which is purely based on word form matching. In addition, NTM labeler boosts the performance of all the methods. For instance, our CGNN model has obtained 17.7\% gains on the F1-score metric with the assistance of extra labeling. Furthermore, our model using CGNN and NTM labeler outperforms all the other baselines. This superior performance indicates that the one-shot learning strategy is competent for our deep QA task with only a small amount of data and the adoption of gated mechanism is also well-suited to work with CNN, adaptively transforming and combining local features detected by the individual filters. 
\begin{table}[t]\centering\small
	
	{
		\begin{tabular}{lcccc}
			\hline
			\hline
			Type  & Proportion  & Precision & Recall & F1-score \\
			\hline
			Background & 16.0  & 38.0 & 56.3 & 45.4 \\
			\hline
			Cause & 8.4  & 48.6 & 67.5 & 56.5\\
			\hline
			Claim  & 4.7  & 47.8 & 65.1 & 55.1\\
			\hline
			Fact  & 51.0  & 23.0 & 42.0  & 29.7\\
			\hline
			Influence & 19.9  & 39.0 & 54.5 & 45.5\\
			\hline
			\hline
		\end{tabular}
	}
	\caption{\label{tab:class} Question type distribution and model performance of CGNN.}
\end{table}
Table \ref{tab:class} shows question type distribution and the performance of CGNN. Our model performs well on most types but drops on \emph{Fact} since these questions are diverse kinds of facts whose patterns are extremely hard to distinguish, even for human. This also shows our task is challenging. 

\paragraph{Model Analysis}

\begin{figure*}
	\centering
	\subfigure{
		\begin{minipage}[b]{0.89\textwidth}
			\includegraphics[width=1\textwidth]{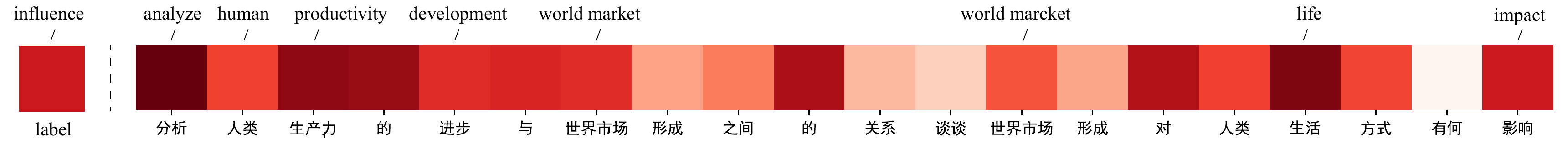}
			\\
			\scriptsize{\emph{Question: please analyze the relationship between human productivity development and world market formation, and evaluate the impact of the formation of global market on human lifestyles.} 
			}
		\end{minipage}
	}
	\subfigure{
		\begin{minipage}[b]{0.89\textwidth}
			\includegraphics[width=1\textwidth]{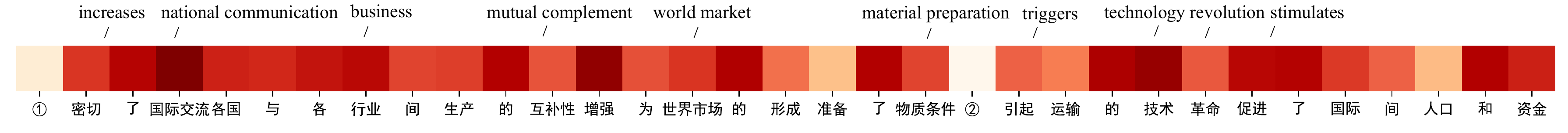}
			\scriptsize{\emph{Answer (extract): the development of human productivity increases the frequency of national communication, strengthens the mutual complement among nations and business, provides sufficient material preparation for global market formation, triggers the technology revolution of transportation and stimulates both migration and liquidity among nations. }
			}
		\end{minipage}
	}
	\caption{Visualization for question and answer representation weights after gated flow. The darker color means the higher weights.} \label{fig:1}
\end{figure*} 

The primary motivation of highway network is to ease gradient-based training of highly deep networks through utilizing gated units. When remaining one gate in our network, our model works similarly to the highway network. To compare different gated mechanisms, we carry out further experiments with the same CNN setting. As shown in Table \ref{tab:AR}, CGNN achieves the best performance in terms of all evaluation metrics, indicating our gated mechanism is well-suited to work with CNN, adaptively transforming and combining local features detected by the individual filters.

In order to give an insight into the effectiveness of the gated mechanism on information flow across neural network layers, we visualize the final weights for each word from question and answer after gated flow as shown in Figure \ref{fig:1}. We can see that the key information (\emph{human, productivity, development, world market, life and influence}) of the lengthy questions and answers will be given higher weights, especially the common words (\emph{world market}) of the question and corresponding answer. Besides, the label in the question is also assigned a high attention, indicating the label works essentially.

\section{Related Work}

\subsection{Question Answering}

Various neural models have been proposed to tackle the tasks of QA task and related knowledge representation \cite{wang2017bilateral,chen2016enhancing,Todor2018Enhancing,kundu2018question}. Previous work mainly focused on factoid questions, falling into the architecture of knowledge base embedding and question answering from knowledge base \cite{khashabi2016question,angeli2015leveraging}. \newcite{Yang:2014} proposed a method that transforms natural questions into their corresponding logical forms and conducted question answering by leveraging semantic associations between lexical representations and knowledge base properties in the latent space. \newcite{Yin:23} proposed an end-to-end neural model that can generate answers to simple factoid questions from a knowledge base. Nonetheless, the knowledge base construction requires a lot of human workload, and the related semantic parsing and knowledge representation will become much more complicated as the scale of the knowledge base increases.

For non-factoid QA, most work formulized it as a semantic matching task on sentence pairs by vector modeling. Different from answer retrieval from knowledge bases, this type of studies used vector to represent QA pairs and compare the distance in vector space to match answer text \cite{Tan:2015,Feng:2015}.  However, the performance of these neural models greatly depends on a large amount of labeled data. 

Our concerned task cannot be simply regarded as either factoid or non-factoid. In fact, questions in our task consist of both factoid and non-factoid ones with purely unstructured corresponding answers. Conventional networks might not be sufficient to represent these QA pairs and we need to seek more powerful models. A recent hot comprehensive QA task is the SQuAD challenge \cite{Rajpurkar2016SQuAD} which aims to find a text span from given paragraph to answer the question. However, our task is quite different from SQuAD. With sufficient learning data, neural models have shown satisfying performance in the SQuAD challenge. In real world practice, examinees are acquired to learn and comprehend meta-knowledge from textbooks so as to pass the exams through review and reasoning among the whole knowledge space instead of simply finding an answer span from a given paragraph, which shows to be challenging for neural networks.

\subsection{Semi-supervised Learning}

For real-world applications of specific domains, the datasets are always inadequate. Semi-supervised Learning methods have been extensively studied to make use of unlabeled data. A batch of models have been proposed based on representation learning \cite{yang2017semi,Lei:2016}. For question answering, current semi-supervised models generally use auto-encoder framework or generative model to obtain the representation of unlabeled data. However, these semi-supervised models can not be applied to generate labels for our raw QA pairs since our task requires diverse types of labels from discourse to sentence which are difficult to annotate, and it is more serious that the entire unlabeled data is also too small to support an effective semi-supervised learning. Thus, we must very carefully choose a proper learning strategy for our task. 

Different from the above research line, one-shot learning belongs to a kind of weakly supervised method \cite{Bordes:2014} that was first proposed to learn information about object categories from one or only a few training images \cite{Li:2006}. Recently, memory-based neural network models have greatly expanded the ways of storing and accessing information. Two promising neural networks with external memory have been proposed to connect deep neural network with one-shot learning. One is Neural Turing Machine \cite{Graves2014Neural}, which can read or write the facts to an external differentiable memory. \newcite{Santoro:2016} proved it to be an effective method for image recognition. The other is Memory Network \cite{Vinyals:2016}. The crucial difference between them is that the latter does not have a mechanism to modify the content of the external memory, which lets us choose the former as our one-shot learning implementation. Compared with previous methods for question answering, our model is more weakly supervised with a limited amount of training data. As to our best knowledge, this is the first attempt that adopts one-shot learning for a deep QA task by using NTM as automatic labeler. 

\section{Conclusion}
This paper presents a neural model with NTM for a challenging deep question answering task from history examinations. Our experimental results show that the adopted CGNN together with other neural models works much better than BM25. With an NTM labeler, all the deep neural models are further enhanced. We also release a Chinese comprehensive deep question answering dataset and have launched a new research line for the Gaokao challenge and solve more complicated questions using deep neural networks to learn semantic representation while the previous work focused on choice questions using simple information retrieval methods \cite{chengtaking}. In fact, open-domain questions that can often be distinguished into different aspects can always benefit from one-shot learnings over a few labeling samples, which has been verified the effectiveness in this paper by only relying on the least task-dependent assumptions.

\bibliography{acl2018}
\bibliographystyle{acl}
\end{document}